\documentclass[conference]{IEEEtran}
\IEEEoverridecommandlockouts
\usepackage{cite}
\usepackage{amsmath,amssymb,amsfonts}
\usepackage{algorithmic}
\usepackage{graphicx}
\usepackage{textcomp}
\usepackage{xcolor}
\def\BibTeX{{\rm B\kern-.05em{\sc i\kern-.025em b}\kern-.08em
    T\kern-.1667em\lower.7ex\hbox{E}\kern-.125emX}}
\begin{document}

\title{Overview Analysis of Recent Development on Self-Driving Electric Vehicles
{\footnotesize \textsuperscript{}}
\thanks{}
}

\author{\IEEEauthorblockN{\textsuperscript{} Qasim Ajao, and Lanre Sadeeq}
\IEEEauthorblockA{\textit{Department of Electrical and Computer Engineering} \\
\textit{Georgia Southern University, Statesboro, USA}\\
}
}

\maketitle

\begin{abstract}
This paper provides a comprehensive overview of recent advancements in autonomous electric vehicle (AEV) within the specified region. It elaborates on the progress and comparative analysis of diverse subsystems, including energy storage, cell balancing for battery systems, vehicle charger layouts, electric vehicle motor mechanisms, and braking systems. Furthermore, this paper showcases several prototype autonomous electric vehicles as conclusive study findings.
\end{abstract}


\section{Introduction}
In recent years, the development of autonomous electric vehicles (AEVs) has gained significant attention from researchers and engineers worldwide. AEVs are expected to revolutionize the way we commute and transport goods, offering safer and more efficient solutions to our transportation needs. Self-driving or autonomous electric vehicles (AEVs) offer numerous benefits including but not limited to the following \cite{b1}:

\begin{itemize}
\item Enhanced Safety: One of the most significant benefits of self-driving vehicles is their potential to increase safety on the road. AEVs are equipped with advanced sensors, cameras, and artificial intelligence technology that allows them to detect and respond to potential hazards more quickly than human drivers.
\item Increased Efficiency: Self-driving cars can optimize their routes and speed, leading to reduced traffic congestion and lower fuel consumption.
\item Improved Accessibility: For individuals with disabilities or limited mobility, self-driving cars can provide greater independence and accessibility.
\item Cost Savings: Self-driving cars could reduce the cost of transportation, as individuals would not need to pay for gas, maintenance, or parking. Additionally, the technology could reduce the cost of shipping goods, as automated trucks could operate more efficiently than human drivers.
\item Environmental Benefits: Self-driving cars can potentially reduce emissions and improve air quality by optimizing their speed, reducing idling time, and using electric powertrains.\\
\end{itemize}

AEVs represent a significant advancement in transportation technology, offering safer, cleaner, and more efficient solutions to our transportation needs. The development of AEVs requires the integration of several key components and subsystems, including the motor system, battery management system, charging and discharging systems, braking system, steering system, and other accessories. Continued research and development in this field will likely lead to further advancements in AEV technology in the future \cite{b2}.

\begin{figure}[htbp]
\centerline{\includegraphics[scale=0.67]{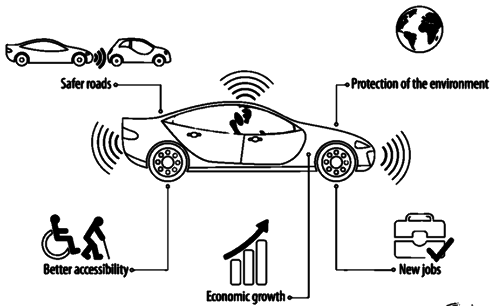}}
\caption{Benefits of Self-Driving in the United States}
\label{fig}
\end{figure}

Self-driving technology has the potential to revolutionize the transportation industry and offer numerous benefits to individuals and society as a whole.

\section{Hybrid AEVs and Plug-in}
AEVs can be classified into three main categories: fully autonomous, semi-autonomous, and connected autonomous vehicles. Fully autonomous vehicles require no human intervention, whereas semi-autonomous vehicles require a driver's assistance in certain situations. Connected autonomous vehicles are equipped with advanced sensors and communication technologies that allow them to interact with other vehicles and the infrastructure \cite{b3}.

Hybrid Autonomous Electric Vehicles (HAEVs) have been widely promoted in the USA over the past decade, and nearly every car manufacturer now has at least one HAEV model in the market. HAEVs were developed to address the energy storage problem faced by electric vehicles at the time. The hybrid system allows for electric power to be obtained from the engine, providing a solution to the battery energy storage issue. HAEVs can be divided into series and parallel hybrids. In the series hybrid, the engine power is entirely connected to the battery, and all motor power is derived from the battery. In contrast, both the engine and motor contribute to the propulsion power in the parallel hybrid, with the torque being the sum of both motor and engine power.\\ 

\begin{figure}[htbp]
\centerline{\includegraphics[scale=0.3]{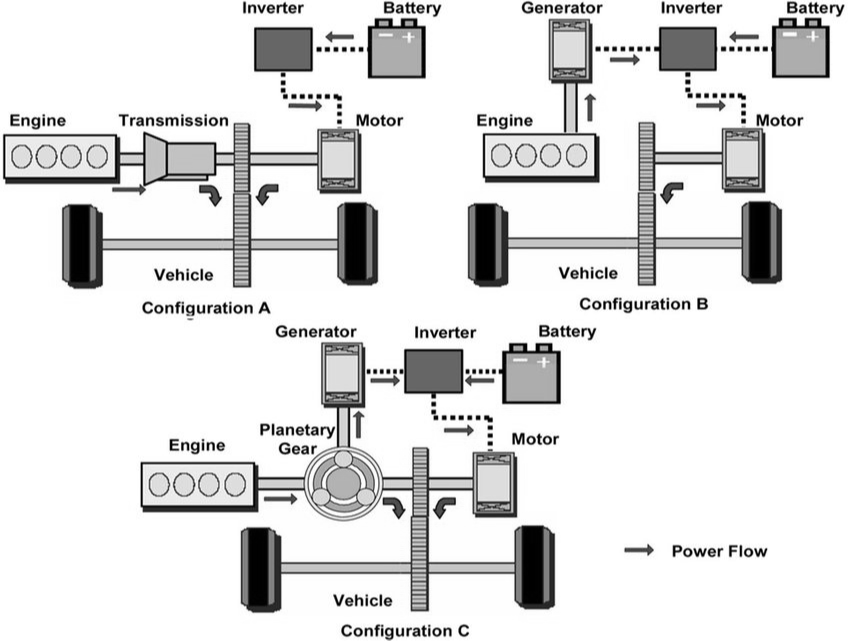}}
\caption{Hybrid vehicle configurations}
\label{fig}
\end{figure}

Fig 2 shows the configuration of a Hybrid vehicle where (A) is parallel, (B) series, and (C) parallel/series which is the power-split. Both series and parallel hybrids can regenerate power during braking or deceleration. However, HAEVs still emit harmful gases, and the introduction of plug-in HEVs has addressed some of these issues. With plug-in HEVs, users can charge the battery using AC power from the mains, providing a convenient way to reduce emissions \cite{b4}.
\begin{figure}[htbp]
\centerline{\includegraphics[scale=0.1]{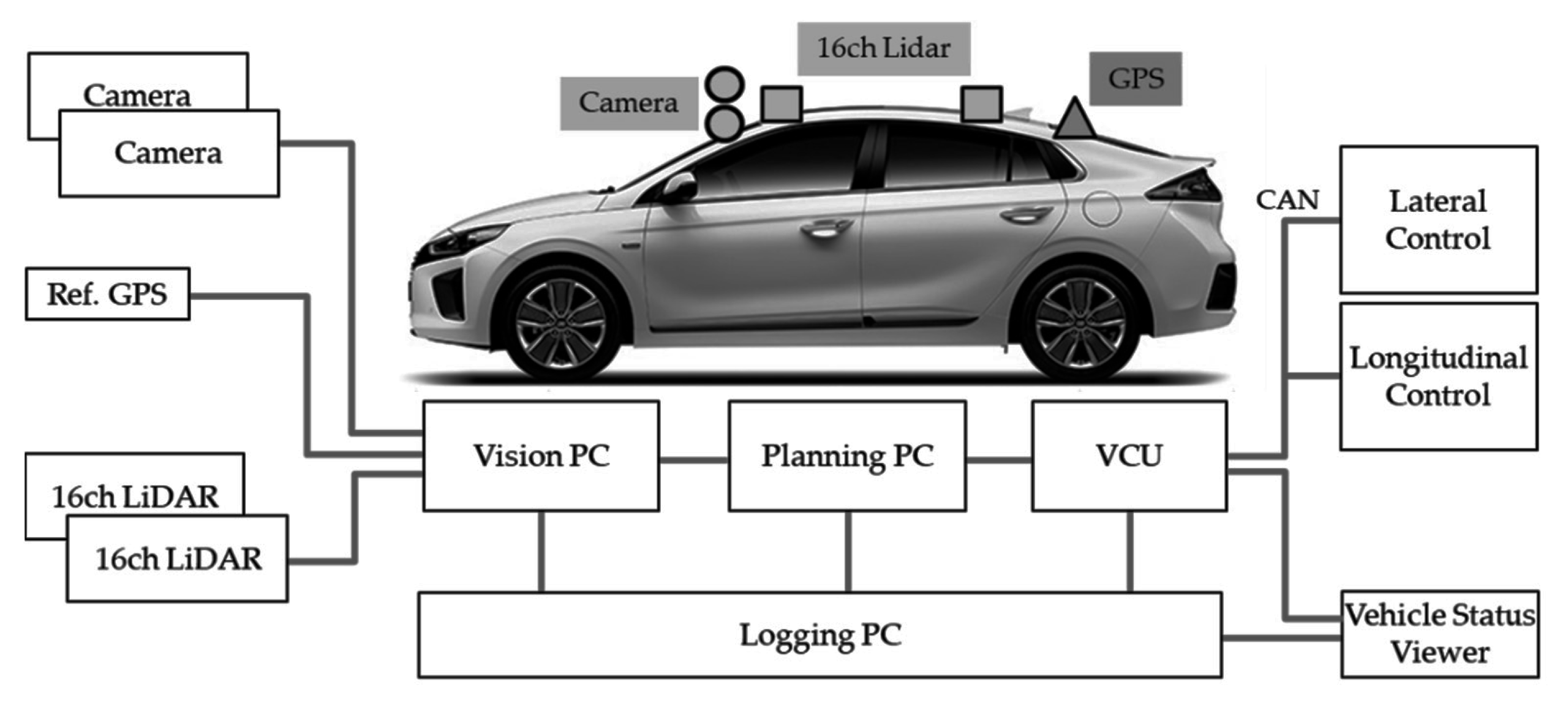}}
\caption{Autonomous EV Hardware Configurations}
\label{fig}
\end{figure}

A test vehicle for autonomous driving is shown in Figure 3. The vehicle was a small sedan, fully electric vehicle. Two cameras and two 16ch LiDAR sensors were used as vehicle sensors. It was also equipped with a high-performance GPS for obtaining reference information. To control the vehicle, control commands were issued via a Controller Area Network (CAN) \cite{b5}.

\section{Components of an AEV}
Autonomous Electric Vehicles (AEVs) utilize an electric propulsion system that completely eliminates the need for internal combustion engines. Electric power serves as the sole energy source for vehicles, leading to highly efficient power conversion through the electric motor's propulsion system. Recent academic and industry reports indicate a significant surge in research and development activities in AEVs. Commercial AEVs are now available in several countries worldwide. Various nations have offered incentives such as lower tax rates or exemptions, free parking, and charging facilities to promote AEV adoption. AEVs are the next frontier in the transportation industry, with their combination of self-driving technology and electric power. They have multiple components that make them unique and technologically advanced.

\subsection{Battery Management Systems (BMS):}
The battery is the primary source of power in an AEV, and the BMS controls the charging and discharging of the battery. The BMS manages the temperature, state of charge, and health of the battery \cite{b1}. This system ensures the battery is functioning optimally, and it provides important information about the battery's condition to the vehicle's control systems \cite{b6}.

\subsection{Electric Motor:}
The electric motor is the powertrain of the AEV. It converts the electrical energy stored in the battery into mechanical energy that drives the wheels. The electric motor in AEVs is typically a brushless DC motor, which is more efficient, quieter, and requires less maintenance than traditional internal combustion engines.
\subsection{Charger and Charging Network:}
The charging system is an essential component of AEVs. It includes the on-board charger that converts the AC power from the charging station to DC power that can be stored in the battery. The charging network comprises the infrastructure of charging stations where AEVs can recharge their batteries. This network is expanding rapidly to support the growing demand for AEVs.
\subsection{Braking System:}
The braking system in AEVs includes regenerative braking and traditional mechanical braking. Regenerative braking converts the kinetic energy generated during braking into electrical energy, which is stored in the battery. This system reduces the wear and tear on the brakes, and it can improve the range of the vehicle.
\subsection{Steering System:}
The steering system in AEVs is electronic, and it is controlled by the vehicle's software. This system allows for precise and responsive steering, and it can be integrated with other systems, such as the autonomous driving system, to provide a seamless driving experience.
\subsection{Capacitors (The Future):}
Capacitors are an emerging technology in AEVs, and they have the potential to revolutionize the industry. Capacitors can store and discharge electrical energy rapidly, which is beneficial in regenerative braking and acceleration. They are also more durable and have a longer lifespan than batteries.

\begin{figure}[htbp]
\centerline{\includegraphics[scale=0.25]{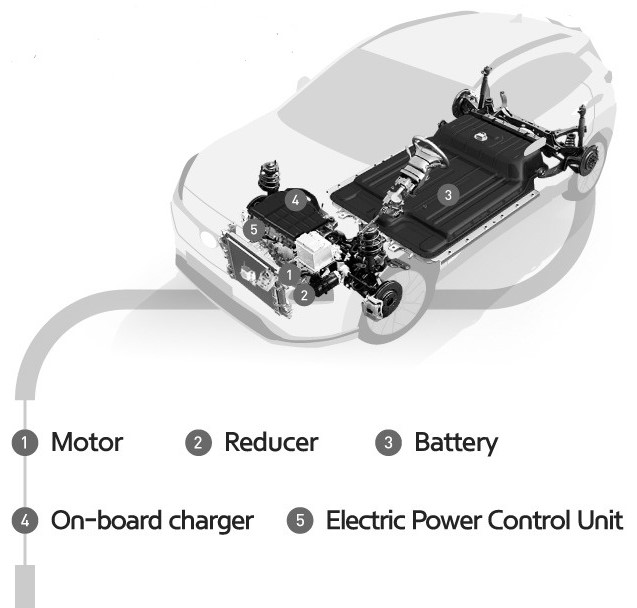}}
\caption{Key Components of an Autonomous Electric Vehicle (AEV)}
\label{fig}
\end{figure}

The key components of AEVs are critical in ensuring their optimal performance, safety, and efficiency. These components are continually evolving and improving, and their development is driving the growth and adoption of AEVs.\\

The Hybrid Autonomous Electric Vehicle (HAEV) is an alternative that has gained widespread usage over the last few years. Virtually all car manufacturers now offer at least one HAEV model. A question arises regarding which type of vehicle will dominate the market and prove suitable for the future. This paper examines recent AEV development and proposes future developments in the field. In the United States, the federal government and some states have provided incentives for the adoption of AEVs, such as tax credits, grants, and other incentives, to promote clean energy and reduce greenhouse gas emissions. The increasing availability of charging infrastructure across the country has also encouraged AEV adoption. Furthermore, many large companies have started to include AEVs in their fleets, further driving their adoption \cite{b7}. 

\section{Energy Storage and Battery Management Systems (BMS)}
The battery is a crucial component of electric vehicles and plays a major role in determining their success. Recently, extensive research and development have been carried out to improve battery technology. Lithium-ion (Li-ion) batteries have emerged as a popular choice for new-generation electric vehicles due to their high energy density and long cycle life \cite{b8}. However, the safety and stability of these batteries have been a concern, particularly for LiCoO\textsubscript{2}, LiMn2O\textsubscript{4}, and Li(Ni\textsubscript{1/3}Mn\textsubscript{1/3}Co\textsubscript{1/3})O\textsubscript{2} types \cite{b2}. The LiFePO\textsubscript{4} type, on the other hand, is preferred for its chemical stability and inherent safety. 

The use of lead-acid batteries is still dominant in low-cost solutions for electric vehicles, which are typically used in applications such as electric wheelchairs, golf carts, micro-cars, and neighborhood town air. NiCd batteries, on the other hand, have been phased out due to environmental concerns, such as the RoHS directive. Table 1 shows the comparison of energy storage systems.

\begin{table}[htbp]
\caption{Energy Storage Systems}
\begin{center}
\begin{tabular}{ |p{2cm}|p{2cm}|p{2cm}|  }
 \hline
 \multicolumn{3}{|c|}{COMPARISON} \\
 \hline
 Storage Systems& Life (Cycles)& Efficiency (\%)\\
 \hline
 Ultra Capacitors   & 10'000-100'000    &93-98  \\
 Flow Batteries&   40 Years & - \\
 Lead Acid &200-500 & 75  \\
 Pumped Hydro    &75 Years & 70-80  \\
 NAS&   2000-3000  & 89\\
 CAES& 40 Years  & 50   \\
\hline
\end{tabular}
\end{center}
\end{table}

The battery system comprises multiple battery cells that are connected in either parallel or series configurations based on the system design. Each of these cells must be closely monitored and regulated for optimal performance. This conditioning monitoring involves measuring the voltage, current, and temperature of each cell. These measured parameters are then utilized to determine the control and protection decision parameters for the system \cite{b9}.

\begin{figure}[htbp]
\centerline{\includegraphics[scale=0.35]{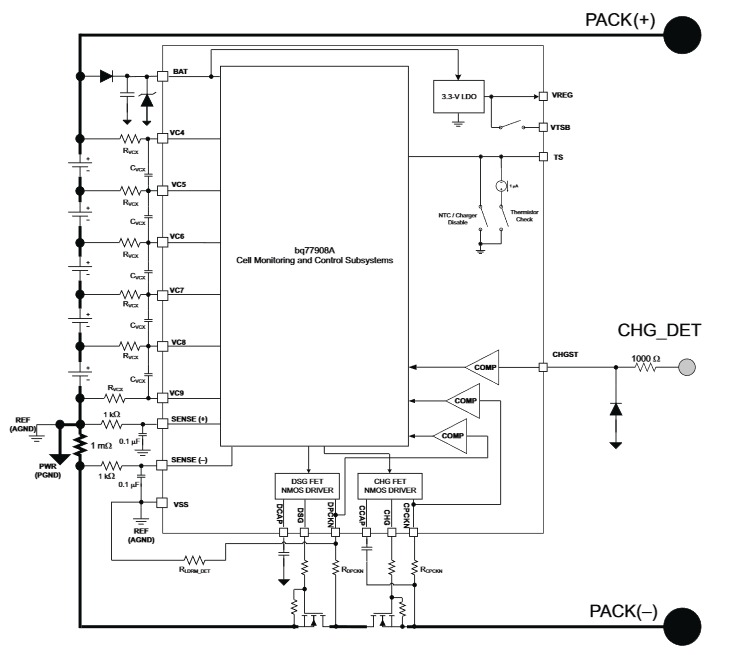}}
\caption{An MCU-independent cell balancer}
\label{fig}
\end{figure}

Typically, two crucial parameters are provided: State of Charge (SoC) and State of Health (SoH). SoC is analogous to the fuel gauge in a car, providing information on the battery's current charge level. It is measured by analyzing the voltage and current information. On the other hand, SoH is used to record the battery's health or aging condition, and there are several definitions for it, although one of the most prominent ones is:

\begin{equation}
SoH =\frac{Nominal\, Capacity-Loss\, of\, Capacity}{Nominal\, Capicity}\
\end{equation}

To avoid overloading a particular cell, the process of cell balancing entails using balancing control to regulate the charging or discharging of each individual cell, thereby ensuring uniform conditioning for all cells \cite{b10}. When working with multiple cells, a balancing system becomes necessary. Basic methods that function without the need for a Micro-controller Unit (MCU) also exist, as depicted in Figure 5.

The articles by Affanni et al. and Chan discuss battery choice and management for new-generation electric vehicles. Affanni et al. emphasize the importance of battery management systems (BMS) to ensure the optimal use of batteries in electric vehicles. They also discuss the different battery technologies and their suitability for various applications. Chan provides a comprehensive review of the present status and future trends of electric vehicles, including battery technology. He highlights the need for further research and development to improve battery performance, cost, and safety \cite{b11}.\\

In addition to battery technology, there have been developments in ultra-capacitors, which can provide rapid charging and discharging capabilities for autonomous vehicles. These devices can store and deliver large amounts of energy in short periods, making them suitable for use in regenerative braking systems and other applications. Ultra-capacitors are also more durable than traditional batteries and have a longer lifespan, making them ideal for use in autonomous vehicles that require a reliable power source. While ultra-capacitors do not have the same energy density as batteries, they can be used in conjunction with batteries to provide a more efficient and reliable power source for autonomous vehicles. However, ultra-capacitors have lower energy density compared to batteries and may not be suitable for long-range electric vehicles. Recently, MIT researchers developed a new technology for batteries using a crystal structure that allows for 100 times faster charging than conventional lithium-ion batteries. This new battery technology could potentially revolutionize the electric car industry, making electric vehicles more practical for daily use by significantly reducing charging times \cite{b12}.\\

Another development is in the discovery of a new lithium iron phosphate electrode that has a unique crystal structure, which allows for rapid charging without degrading the battery's lifespan. With this breakthrough technology, it would only take a few minutes to charge an electric car battery to full capacity, instead of the hours it takes with conventional lithium-ion batteries. This technology also has the potential to reduce the size and weight of batteries, making electric cars lighter and more energy efficient In addition to fast charging, another alternative technology that has been gaining attention in the autonomous car industry is ultra-capacitors \cite{b13}. Ultra-capacitors, also known as super-capacitors, can store and discharge energy very quickly, making them ideal for applications that require fast charging and discharging.

\section{Battery Cell Balancing for Improved Performance in AEVs}
\begin{figure}[htbp]
\centerline{\includegraphics[scale=0.36]{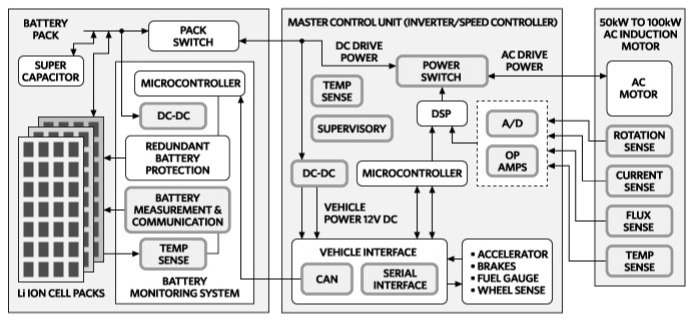}}
\caption{BMS Role in Modern AEV Propulsion Systems}
\label{fig}
\end{figure}

To extend the range and service life of electric vehicles and portable electronic products, it is important to have techniques that balance the charge/discharge characteristics of a battery's individual cells. Passive cell balancing technologies, which use impedance tracking, coulomb counting, and other state-of-charge monitoring techniques, are already used in the protection and management systems of most 5-20 kWh batteries in hybrid-electric and plug-in hybrids. However, these systems can only unlock a portion of a battery's "stranded" capacity as they allow cells with higher capacity to repeatedly bleed off energy from weaker cells. This raises concerns about whether passive balancing can be used for 25-100 kWh batteries in pure EVs \cite{b2}. 

The better alternative is active cell balancing, which redistributes the charge from stronger cells to weaker ones within a battery stack. While electronics manufacturers are promoting the benefits of their early active balancing technologies, battery and electric vehicle manufacturers have concerns about the added cost and complexity of these systems. Passive balancing systems, which are considered "smart," utilize impedance tracking, coulomb counting, and other techniques to monitor the state-of-charge (SoC) as shown in Figure 5.

\subsection{Cell Uniqueness and Qualities}\label{AA}
Just like snowflakes, each Li-ion cell has its own unique blend of anode, cathode, and electrolyte materials, which can vary significantly among manufacturers. The resulting cell voltages typically range from 2.7 V to 4.25 V, with lithium-phosphate cells used in many EVs and HEVs having a voltage of around 3.6 V. Even cells with the same chemistry have some level of variation in capacity, open-circuit voltage, charge capacity, self-discharge rate, impedance, and thermal characteristics that can impact their SOC. Additionally, as the battery ages, the differences between cells become more pronounced, resulting in increased capacity and service life degradation as shown in Figure 1.

\begin{figure}[htbp]
\centerline{\includegraphics[scale=0.4]{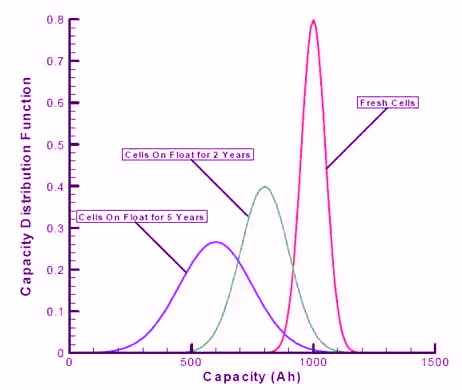}}
\caption{Battery Capacity Vs. Function}
\label{fig}
\end{figure}

When cells are connected in series, the stack can only be charged to the weakest cell's capacity, even if other cells can accept a deeper charge. Lithium-based chemistries are highly reactive, and pushing additional current into the stack can cause over-voltage conditions in fully charged cells, leading to heat damage and, in extreme cases, fire. A cell's charge must be controlled within 10-50 mV to avoid thermal runaway, and a minimum resolution of 2-5 mV is necessary to prevent damage that could reduce the cell's capacity and service life. The weakest cell also governs the stack's current delivery capacity, preventing it from being depleted to the point where dendrites, or crystal structures, precipitate from the cell's electrolyte, leading to micro-shorts within its structure \cite{b6}.

\subsection{BMS Implementation Strategies}\label{AA}
All battery equalization methods must operate within the constraints of the battery pack's remaining battery management and protection features. 

\begin{figure}[htbp]
\centerline{\includegraphics[scale=0.51]{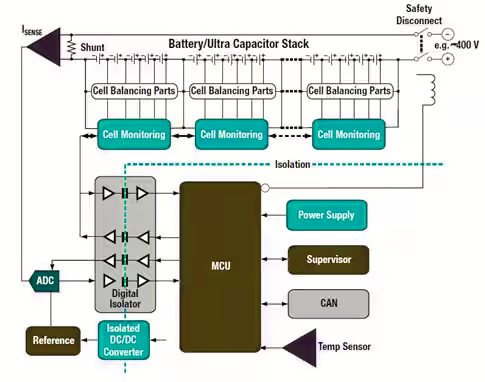}}
\caption{A block diagram of a typical BMS}
\label{fig}
\end{figure}

In the majority of automotive designs, an automotive-qualified host MCU is used to execute the software for cell balancing algorithms and control functions, usually located inside the battery management system (BMS) itself (as shown in Figure 6). The BMS MCU can frequently employ the same electronics used to calculate cell voltage, charge/discharge current, and state of charge (SOC) for the battery's "fuel gauge" and charge management systems to perform similar measurements required for its cell balancing operations \cite{b14}.

A precise voltage measurement or a combination of coulomb counting and another technique that calculates the total current flowing in and out of the cell can be used to determine a battery cell's SOC. In both situations, an A/D converter with 12 to 14-bit resolution is required for cell voltage measurement. While most producers do not provide complete integrated solutions for active cell balancing, off-the-shelf products are available that handle essential A/D, multiplexing, level shifting, and communication functions, which can decrease their component count, module size, and BOM cost.

\subsection{Active vs. Passive cell balancing}\label{AA}
To access more of a battery's available capacity, it is necessary to incorporate circuitry that equalizes the charge between cells \cite{b2}. Current designs mostly use passive balancing techniques, which utilize resistors to discharge or "top balance" overcharged cells, causing their output voltage to drop below the charger's voltage-regulation point, allowing the rest of the stack to continue charging \cite{b6}.

\begin{figure}[htbp]
\centerline{\includegraphics[scale=0.5]{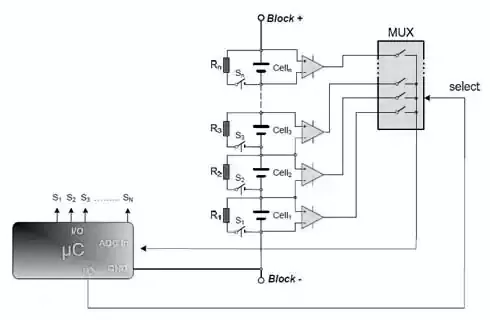}}
\caption{Cell Balancing}
\label{fig}
\end{figure}

Passive balancing is generally only performed during charging because it depletes battery energy. To prevent damage, the individual cells must be frequently sampled, typically at a rate of 4-10 sps for laptop batteries and 20-100 sps for AEV/HEAV batteries, depending on the battery's chemistry and maximum charge rate. Passive balancing techniques are simple to implement and cost-effective, but they suffer from energy losses and thermal issues. Passive cell balancing can be achieved using individual components and a micro-controller unit (MCU), as illustrated in Figure 9, or by utilizing one of the more consolidated options provided by various manufacturers. Active balancing techniques, on the other hand, employ inductive charge transfer to move charge from cells with higher SOC to weaker cells, thereby extracting more usable energy and reducing damage to weaker cells during normal operation. Active balancing can result in up to 20\% more charge cycles, but the higher cost and complexity of the circuitry currently limit its popularity. Thus, this article focuses on passive balancing techniques.

\subsection{Charger Layout}\label{AA}
A passivity-based control method for a phase-shifted resonant converter is famous and regularly adopted, which is a type of power converter commonly used in battery charging systems for autonomous electric vehicles (EVs) \cite{b1}. Both slow and fast charging of AEV batteries requires high-power charging systems, which can be achieved using an H-bridge power converter.

\begin{figure}[htbp]
\centerline{\includegraphics[scale=0.7]{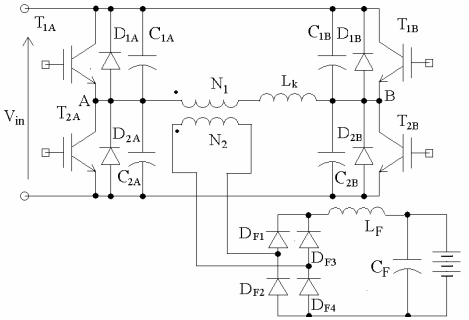}}
\caption{A H-bridge converter}
\label{fig}
\end{figure}
The H-bridge power converter is a well-known type of power converter that is widely used in various applications, including battery chargers and DC-DC converters. It is known for its high efficiency and ability to handle high power. However, the control of H-bridge power converters can be challenging, particularly when operating in a resonant mode.

To address this issue, a passivity-based control method is proposed for the phase-shifted resonant converter. This approach uses the concept of passivity, which ensures the stability and robustness of the system, even in the presence of uncertainties and disturbances. The control method is based on a Lyapunov function, which is used to derive the control law for the resonant converter. The effectiveness of this proposed method is demonstrated through simulations and experimental results. The passivity-based control method can achieve a fast transient response and stable operation, even when operating in the resonant mode. This method also provides good performance under varying load conditions and is robust to parameter variations.

\subsection{Network and Fast Charging}\label{AA}
The charging of autonomous electric vehicles (AEVs) is critical due to the uncertainty of power needs, location, and charging time. The lead-acid battery technology restricts charging rates to less than 0.2C, while Li-ion batteries can be charged at a recommended rate of 0.5C. EVs typically have an on-board battery charger, and charging stations require a suitable transaction program to calculate the tariff. The power needed for charging stations is usually not a concern, with a standard charging power of less than 2.8kW for private cars \cite{b15}. Charging an EV every 3 days in Hong Kong would only affect power consumption by less than 2\%.\\

To enable fast charging of electric vehicles, a high current is required, which is typically facilitated using three-phase power. However, the use of three-phase sockets may not be feasible for all users. To address this, alternative methods have been proposed:

\begin{itemize}
\item Magnetic contactless charging that uses magnetic induction to transfer power without metal contacts.
\item Another option is to use high voltage power transfer, where the power source is stepped up to several kV, allowing for smaller cables to be used.
\item The third option is battery rental, where users swap their batteries at charging stations, reducing the time needed for charging. The EV design must be compatible with this arrangement. Charging stations can also use the vehicle battery charging for energy storage to manage peak demand.
\end{itemize}

\subsection{Anti-lock System and Power Regeneration}\label{AA}
In the past, mechanical braking systems, such as disc brakes, were used for vehicles. However, the braking system for autonomous electric vehicles incorporates both mechanical and electrical braking methods. Initially, during the early stages of braking, electrical power regeneration braking was utilized. This is done to recover kinetic energy from the vehicle during deceleration or while driving down a slope, and return it to the battery. During the final stages of braking, mechanical braking is used to strike a balance between energy conservation and safety.

\begin{figure}[htbp]
\centerline{\includegraphics[scale=0.3]{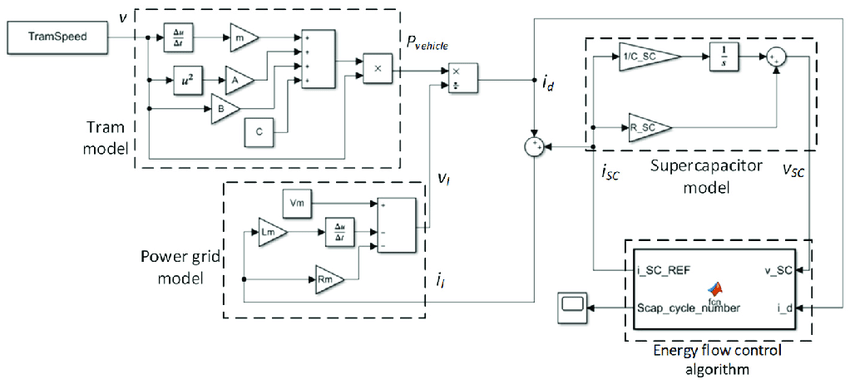}}
\caption{Simulation Model of the Regenerative Braking System}
\label{fig}
\end{figure}

Today, we have the ability to produce motors with high power regeneration. However, a compromise must be made between motor size, weight, cost, power regeneration efficiency, and safety. To expand the range of power regeneration, the motor must be designed to accept high power regeneration modes, which provide high reverse torque to halt the vehicle. In addition, the motor drive should have a high-frequency decoupling capacitor to absorb the rapid transients of the reverse current. Figure 10 shows the Simulation Model of the Regenerative Braking System an autonomous EV in MATLAB/Simulink schematic \cite{b16}.

Most vehicles have conventional anti-lock braking systems (ABS) installed to prevent skidding and achieve stable braking performance. The braking characteristics depend on the wheel slip and road conditions. Continuous slip changes and discrete valve actions are combined to create a discrete hydraulic pressure, and PID and finite state machine theories are utilized in the anti-lock braking system. ABS optimization is intended to maximize tire forces under any road conditions. Thus, the system must identify the wheel slip ratio that corresponds to the peak tire-road adhesion characteristic. These peak values vary widely depending on the road, tire, and other factors. For any rolling conditions, the optimal wheel slip rate will be utilized as a control reference to optimize braking force. 

\section{Future Development}
In 2022, several car manufacturers and technology companies have continued to make progress in the development of autonomous electric cars. One notable development is the advancement of artificial intelligence and machine learning algorithms used in autonomous driving technology. These algorithms allow cars to make more accurate and reliable decisions in complex driving scenarios.

\begin{figure}[htbp]
\centerline{\includegraphics[scale=0.28]{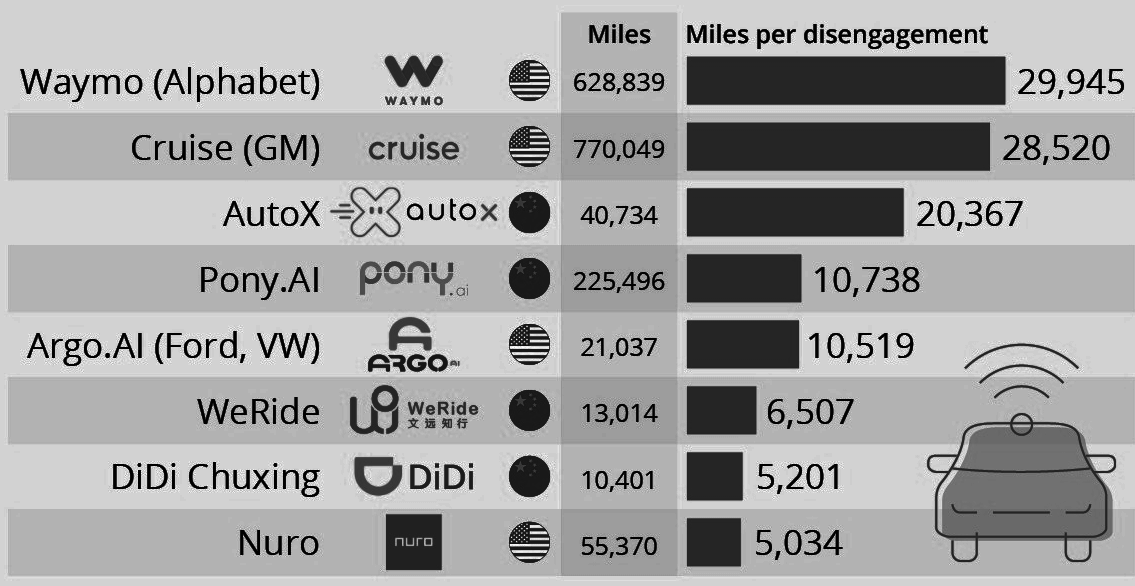}}
\caption{Self-Driving Manufacturers Going the Distance}
\label{fig}
\end{figure}

Another important development is the integration of sensors and other advanced technologies, such as lidar and radar, to provide a more complete view of the car's environment. This helps to improve the car's ability to detect and avoid obstacles and other hazards on the road.

As of April 2023, there have been several developments and discoveries on Autonomous Electric Vehicles (AEVs) in 2022. One of the key advancements has been the improvement of electronic parts and accessories for AEVs, particularly in the areas of propulsion, safety, and control. Many local and overseas companies and institutions have been working on self-driving electric vehicles, including Tesla, Google, and Apple. Tesla has been working on its Autopilot system, which allows for semi-autonomous driving, while Google has been developing its Waymo self-driving technology. Apple has also reportedly been working on its own autonomous car project, although details are scarce. In terms of propulsion, there has been progress in the development of new battery technologies, such as solid-state batteries, which promise higher energy density, longer range, and faster charging times than current lithium-ion batteries. Companies such as Toyota, Samsung, and QuantumScape are among those investing in solid-state batteries.\\

In the area of safety, there have been advancements in sensors and software that enable AEVs to better detect and respond to their surroundings. Lidar (Light Detection and Ranging) sensors, in particular, have become more common in AEVs, providing detailed 3D mapping of the environment around the vehicle. Meanwhile, in the area of control, there has been progress in developing machine learning algorithms that enable AECs to better navigate complex and dynamic environments. For example, NVIDIA has been developing an AI platform called DRIVE that provides AECs with the ability to learn from their surroundings and adapt to changing conditions in real-time.

Despite significant progress in autonomous vehicle technology in recent years, it is still far from perfect. Many companies are testing their electric vehicles in California to enhance software capabilities and safety. Disengagements are a critical part of the testing process and occur when a vehicle's software detects a failure or a driver perceives a failure, leading to control being taken over.\\

According to data published by The Last License Holder website, 29 companies reported that they were actively testing self-driving cars on public roads in California. In total, they covered 1,955,208 miles in autonomous mode, with 3,695 disengagements occurring. Google's Waymo is far ahead of the competition in terms of flawless autonomy, with its fleet covering 630,000 miles last year with 29,425 miles per disengagement. This performance represents a significant improvement over the previous year's 13,219 miles per disengagement. The results in 2020 were much better than those of many companies attempting to perfect the technology, particularly heavyweight competitors like Apple. Last year, Apple's vehicles drove 18,805 test miles with just 144.6 miles covered per disengagement in comparison.

Advancements in EV infrastructure, battery technology, and autonomous freight and cars as marketplaces are expected in 2023. The developments will be driven by the development of an electrification ecosystem. Battery technology advancements that allow consumers and businesses to drive farther on fast charges will be the centerpiece of the development. Autonomous vehicle advancements are shifting from passenger cars to commercial vehicles, with self-driving heavy trucks leading the way. A global shortage of truck drivers and other logistics workers will force companies to turn to technology to address these issues.\\

The Volkswagen ID7, formerly known as the ID Aero, is set to hit the U.S. market next year with VW's best battery range and will be imported from Germany. Qualcomm has launched a new chip that integrates assisted-driving and entertainment functions to reduce costs. Self-driving truck company Gatik and tire maker Goodyear have collaborated to develop tire technology that feeds road condition data to autonomous systems in Canada. Valeo has introduced new safety technology to improve the anticipation of pedestrians, cyclists, and other vulnerable road users. Meanwhile, Chrysler plans to be the first Stellantis brand in North America to roll out new cockpit technology. Additionally, Bosch showcased its smart camera safety system.

\begin{figure}[htbp]
\centerline{\includegraphics[scale=0.23]{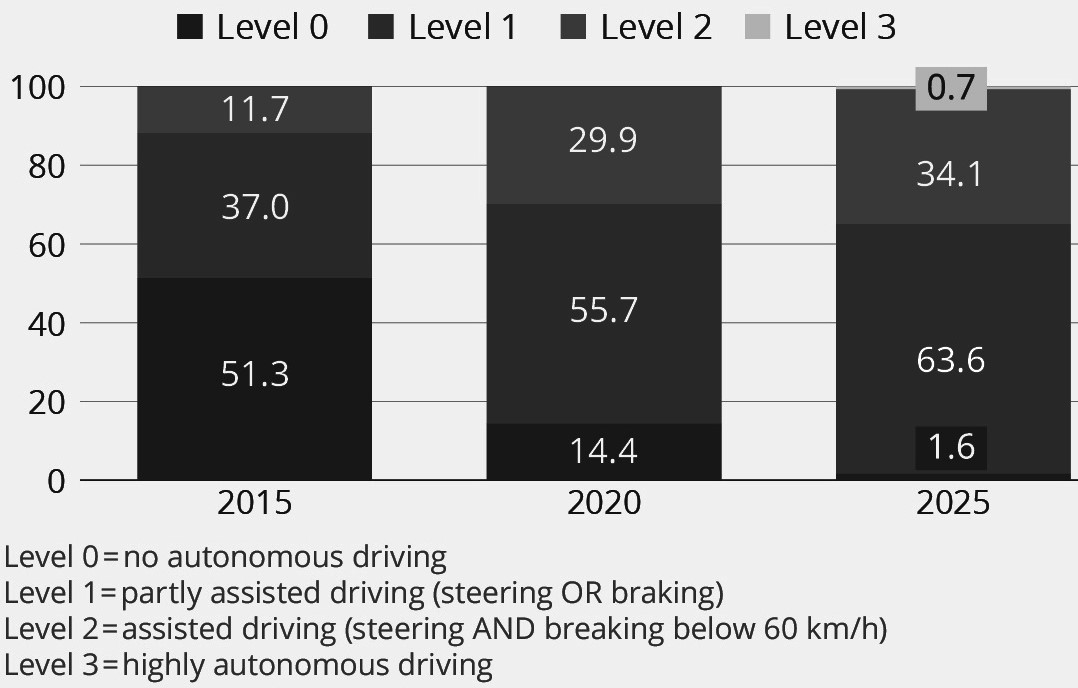}}
\caption{Simulation Model of the Regenerative Braking System}
\label{fig}
\end{figure}

SAE International, the standards developer, has classified passenger car automation levels into six different categories, with levels 0 to 5 representing the autonomous driving ability. The chart includes four of these levels. Levels 1 and 2 fall under the category of assisted mode where the autonomous driving functions assist the driver. Production for level 3 automated driving provisions is now becoming regular, while level 4 is expected to be available by 2025.

Level 3 is the entry-level for an automated mode where the vehicle can drive independently to some degree, but the driver is still required to take over upon request and with advance notice. On the other hand, from level 4, vehicles can drive independently without the driver having to take over. In levels 1 and 2, the autonomous driving systems can assist the driver, but the driver cannot divert their attention from the road. Level 1 provides assistance either for braking or steering, while level 2 offers a combination of both, equipping the car with lane centering and adaptive cruise control capabilities \cite{b17}\cite{b18}.

\section{Conclusion}
The focus of this paper is on the progress made in the development of Autonomous Electric Vehicles (AEVs). It begins by outlining the fundamental structure of AEVs and covers topics such as energy storage, battery management, and vehicle braking systems. The paper also provides insights into the future components of AEVs, economic plans, and strategic developments. Additionally, the paper presents an overview analysis of the recent advancements in AEV technology within the region.

\vspace{12pt}
\color{red}

\end{document}